\documentclass[10pt,twocolumn,letterpaper]{article}

\usepackage{iccv}
\usepackage{times}
\usepackage{epsfig}
\usepackage{graphicx}
\usepackage{amsmath}
\usepackage{amssymb}

\usepackage{comment}
\usepackage{color}
\usepackage{cancel}
\usepackage{epsfig}
\usepackage{booktabs}
\usepackage{algorithm}
\usepackage{algorithmic}
\usepackage{amsfonts}
\usepackage{bm}
\usepackage{float}
\usepackage{subfig}
\usepackage{multirow}
\usepackage{epigraph}
\usepackage{wrapfig}
\usepackage[font=footnotesize,labelfont=footnotesize]{caption}
\usepackage{makecell}
\usepackage{colortbl}
\usepackage{color}
\usepackage[normalem]{ulem}
\definecolor{gg}{rgb}{0.8,0.8,0.8}
\usepackage{xspace}
\usepackage{xr-hyper}
\usepackage[sort&compress,numbers,square,comma]{natbib}

\usepackage{hyperref}
\hypersetup{
    colorlinks=true, %
    linktoc=all,     %
    linkcolor=blue,  %
}

\iccvfinalcopy %

\def\iccvPaperID{9111} %

\ificcvfinal\fi
\newcommand{\mw}{$\mathcal{M}_w$\xspace}
\newcommand{\ms}{$\mathcal{M}_s$\xspace}
\newcommand{\da}{$\mathcal{D}_1$\xspace}
\newcommand{\db}{$\mathcal{D}_2$\xspace}

\begin{document}

\title{The Pursuit of Knowledge: \\Discovering and Localizing Novel Categories using Dual Memory}

\author{Sai Saketh Rambhatla$^{1}$\qquad Rama Chellappa$^2$\qquad Abhinav~Shrivastava$^1$\\[0.3em]
  \hfill$^1$University of Maryland, College Park\quad\hfill
  $^2$Johns Hopkins University\hfill
}

\maketitle
\ificcvfinal\thispagestyle{empty}\fi

\begin{abstract}

We tackle object category discovery, which is the problem of discovering and localizing novel objects in a large unlabeled dataset. While existing methods show results on datasets with less cluttered scenes and fewer object instances per image, we present our results on the challenging COCO dataset. Moreover, we argue that, rather than discovering new categories from scratch, discovery algorithms can benefit from identifying what is already known and focusing their attention on the unknown. We propose a method that exploits prior knowledge about certain object types to discover new categories by leveraging two memory modules, namely Working and Semantic memory. We show the performance of our detector on the COCO minival dataset to demonstrate its in-the-wild capabilities.

\end{abstract}

\section{Introduction} \label{sec:intro}

Unsupervised visual category discovery aims to automatically identify recurring ``patterns,'' which can be objects or object parts, and consequently learn recognition models to identify them from a large collection of unlabeled images with minimal human supervision. 
Such algorithms can dramatically reduce annotation costs as they only need labels for already mined object clusters.
Moreover, automatically discovering categories from unstructured data can help mitigate the biases that occur when manually constructing datasets by collecting images for a fixed-set of concepts. 
Then, why is this setup not a mainstay in recognition?

\begin{figure}
    \centering
    \includegraphics[width=\linewidth]{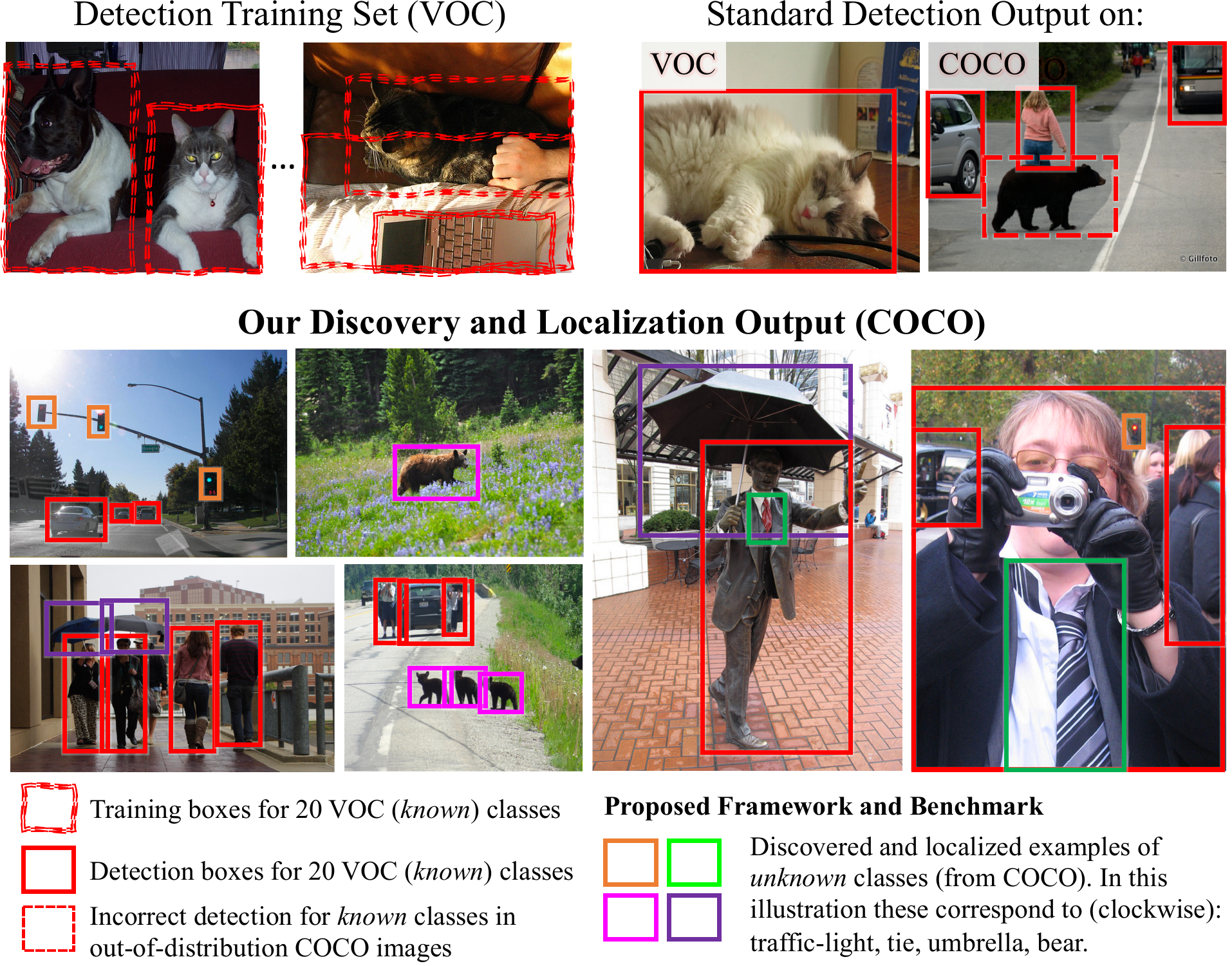}
    \vspace{-0.22in}
    \caption{\textbf{Motivation}. A detector trained on 20 VOC (\emph{known}) classes struggles on out-of-distribution images (\eg, COCO) in the presence of novel (\emph{unknown}) objects (\eg, bear). Our discovery and localization framework builds on this detector and can reliably localize and group semantically meaningful ``patterns" in images with both known and novel objects in challenging images. Novel objects belonging to the same class are assigned the same bounding box color. Best viewed in color.}
    \vspace{-0.2in}
    \label{fig:my_label}
\end{figure}

A key issue is the lack of consistent task definition, with two disparate families of approaches aiming to address unsupervised object discovery.  The first set of approaches~\cite{Vo20rOSD,VoBCHLPP19,hal-01110036,hal-01153017} reduces the problem to co-localization or co-segmentation, by only answering whether a pair of images share the same object and then localizing them. The other set of, arguably more generic, approaches~\cite{5540237,LeeG11,1541280} use clustering techniques to discover semantically similar regions and can sometimes leverage prior knowledge. Because of these different goals, each set defines evaluation protocols that are incompatible with others. The second issue that limits the wide adoption of unsupervised discovery is the scalability of contemporary approaches -- most works use small curated datasets and cannot scale to a realistic setup with a large number of images, a large number of object categories, and complex images (\eg, a large number of objects and categories per image). Finally, a desirable property of unsupervised discovery approaches, often lacking in contemporary works, is online processing of new data as it arrives. To address these issues, it is necessary to define a standardized protocol (datasets and metrics), that more closely reflect real-world requirements of a task like unsupervised object discovery. 

Towards this, this paper proposes a large-scale benchmark for evaluating unsupervised object discovery approaches and a scalable never-ending discovery approach that can deal with real-world complexities. Our approach is loosely inspired by how humans learn -- continuously, utilizing prior knowledge. Our benchmark is designed to evaluate such continuous and knowledge-driven learning and reflect real-world complexities and scale, and as a practical consideration, it is amenable to using available pretraining datasets in object recognition literature.

\smallskip\noindent\textbf{\underline{Motivation}:} Humans never stop learning~\cite{1714515, Lillard2011OldDL}.  Since infancy, we continuously stumble upon objects that we have never encountered before and learn to recognize them with time~\cite{Johnson10568}. This process is mostly reliant on the current knowledge we possess. For example, a toddler, with a pet dog at home, may point to a lion at a zoo and mistakenly call it a dog; but the toddler is unlikely to point to a bench, because she hasn't learned about chairs and couches yet~\cite{409}. This alludes to a continuous learning paradigm, where we notice new objects, associate them with our current knowledge~\cite{Johnson2010HowIL}, and update our knowledge (learn new concepts or update existing concepts). So how are we able to do this so effortlessly, and often unconsciously?

Studies have demonstrated that ``memory'' is the primary form of representation that brings together perception and learning~\cite{3233801}, and knowledge, categories, \etc, all derive from memory~\cite{article}. Despite overwhelming evidence of its importance, memory is arguably one of the least studied components in computational visual recognition; its role often ceded to other modules; \eg, pretrained weights of a neural network~\cite{liang2015towards}, forms of knowledge-graphs~\cite{marino2016more,Wang_zslCVPR2018}. In this work, we investigate how memory can be used to represent knowledge and drive the discovery of new concepts and develop a continuous learning approach that is inspired by well established memory processes~\cite{ATKINSON196889}.

\smallskip\noindent\textbf{\underline{Contributions}:} Our two complimentary contributions are: a scalable concept discovery and localization approach, and a realistic protocol for evaluating discovery approaches.

\textbf{Our first contribution}, 
a never-ending concept discovery and localization framework, is built using two memory modules: Semantic and Working memory, jointly referred to as \textbf{dual memory}. Semantic memory is the portion of long-term memory that contains concepts from past experience (current knowledge).
Working memory, on the other hand, is synonymous to short-term memory and is responsible for accumulating and temporarily holding information for processing. 
Our final algorithm consists of carefully crafted operations which compare a new region to the dual memory, decide if it is known (or already discovered) or a novel category, and accordingly update the relevant memory modules. When sufficient concepts accumulate in the working memory, our algorithm amalgamates learned concepts from the working memory into the semantic memory, which then become part of prior knowledge. This process can continue in a never-ending, online fashion.

\textbf{Our second contribution} is a simple, realistic benchmark for never-ending concept discovery, not only to evaluate our approach, but also to enable future works to compare on a standardized protocol. 
We argue that most standard benchmarks are either not realistic, are designed for classification, are suited more for co-segmentation than discovery, are not labeled to evaluate performance, or are too small in scale. 
In addition, earlier works on never-ending and large-scale learning~\cite{shrivastava_eccv12,Chen-2013-7815,CVPR.2014.412} either assume an a-priori list of concepts, which is unrealistic, utilizes crawled internet images which is not reproducible, or use proprietary datasets which cannot be released, ruling out future comparisons.
Therefore, we propose a simple, realistic benchmark for never-ending concept discovery and localization. 

Our benchmark is designed to leverage three existing datasets, ImageNet, Pascal VOC, and COCO. We assume ImageNet and VOC are the labeled datasets available for pretraining and detecting 20 classes, respectively, from which prior knowledge can be derived; and  COCO is the in-the-wild dataset to perform concept discovery and localization. The 20 VOC classes are treated as \emph{known} classes, and the 60 additional classes in COCO are \emph{unknown} classes used to evaluate a subset of the discovered objects. Other discovered objects are qualitatively assessed. This setup has six desirable properties: (1) these datasets are widely used, and therefore, we know the performance if all 80 classes were labeled; (2) the discovery set consists of all known and a variety of unknown classes; (3) the images in COCO have a slightly different distribution from VOC; (4) dataset is large-scale; (5) bounding box labels are available and (6) future progress on these datasets~\cite{GuptaDG19} can be leveraged.

\vspace{-0.05in}
\section{Related Works}
\vspace{-0.05in}

\noindent\textbf{Similar learning paradigms.} Several paradigms have been proposed to imitate continuous learning models, including, but not limited to, incremental learning~\cite{9009019,RebuffiKSL17,CastroMGSA18,Dhar_CVPR19,ShmelkovSA17}, never-ending learning~\cite{Chen-2013-7815,CVPR.2014.412}, open-world learning~\cite{Bendale_2015_CVPR}, semi-supervised~\cite{shrivastava_eccv12,NIPS2014_5352,videossl_cvpr15}, and omni-supervised learning~\cite{RadosavovicDGGH18}. Each of these paradigms is actively pursued by computer vision and machine learning researchers; and an equitable and comprehensive discussion with even the relevant approaches is unrealizable given the manuscript length constraints. We discuss the differences between these paradigms and the problem setup for our approach. %

Most approaches mentioned above differ from ours in one of the following aspects: (a) our focus is the localization task, unlike most incremental~\cite{9009019,RebuffiKSL17,CastroMGSA18,Dhar_CVPR19}, open-world~\cite{Bendale_2015_CVPR}, and semi-supervised learning approaches~\cite{shrivastava_eccv12};  (b) we do not assume any a-priori list of concepts of interest (novel classes), unlike most incremental~\cite{9009019,RebuffiKSL17,CastroMGSA18,Dhar_CVPR19}, semi-supervised~\cite{shrivastava_eccv12}, and never-ending learning frameworks~\cite{Chen-2013-7815,CVPR.2014.412}; (c) unlike contemporary discovery methods~\cite{hal-01110036,hal-01153017,VoBCHLPP19, Vo20rOSD}, we operate in a real-world setup, where images contain many objects, including ones that are similar to known objects; (d) scalability to larger datasets and objects. While our work shares the goal of \cite{Chen-2013-7815} in learning models for never-ending learning for training object detectors, there is one technical and several practical differences. \textbf{First}, \cite{Chen-2013-7815} is \emph{language-driven} (ref.\ $\S$1.1c in \cite{Chen-2013-7815}), \ie, it starts with an a-priori list of concepts and then searches images for that concept, and then attempts to build the detectors. As opposed to this, our approach does not start with any a-priori list, and discovers in the set of images it is given. We believe this is a crucial distinction. \textbf{Second}, \cite{Chen-2013-7815} assumes at least a few `canonical' images for concepts and uses it to train detectors (cf.\ $\S$3.1 in \cite{Chen-2013-7815}). In contrast, our approach can discover and localize objects directly in complex images. \textbf{Finally}, data used by \cite{Chen-2013-7815} has not been publicly released making it impossible to compare with it fairly on the same set of images. Moreover, the released code relies on the discussed assumptions, making it unsuitable for the current setup (COCO dataset).

\noindent\textbf{Object Category Discovery.} Category discovery is the problem of identifying semantically similar recurring patterns in unlabelled data. Object category discovery works can be broadly divided into two categories: image~\cite{HsuLSOK19,HsuLK16,han20automatically,han19learning,HsuLK18,singh-cvpr2019} and region-based approaches~\cite{VoBCHLPP19,5540237,LeeG11,doersch2014context,6126314,Osep19ICRA,Shaban19A,8365833,xie2019object,hal-01153017,hal-01110036,5995527,Karpathy_ICRA2013,Romea-2014-17150,GalleguillosML14,Arandjelovic19,1541280,discover_seg_cvpr14}. Our discovery framework is a region-based approach. Recent region-based methods~\cite{hal-01153017,hal-01110036,VoBCHLPP19,Vo20rOSD}, do not assume any prior knowledge and solve an optimization problem to discover new objects. Unlike such methods, we assume a set of known objects and leverage this knowledge to discover novel objects, similar to~\cite{5540237,LeeG11}. Lee \etal~\cite{5540237} proposed object graphs to incorporate context which improves clustering for the discovery of novel categories. In contrast, we use cluster prototypes to improve clustering; and unlike~\cite{5540237}, we do not require the number of clusters as an input. Improving on their previous work~\cite{5540237}, Lee~\etal~\cite{LeeG11} proposed an iterative, self-paced approach for category discovery that focuses on the easiest instances first, and then progressively expands its repertoire to include more complex objects. Our approach is implicitly self-paced and does not need a specific curriculum definition for discovery. 

For an in-depth review of other object discovery methods, memory formulations, and our relationship to learning paradigms (\eg, supervised, incremental, open-world, never-ending), kindly refer to the \textbf{supplementary} material.

\section{Approach Overview}
\label{sec:exp}

\begin{figure}[t]
    \centering
    \vspace{-0.05in}
    \includegraphics[width=\linewidth]{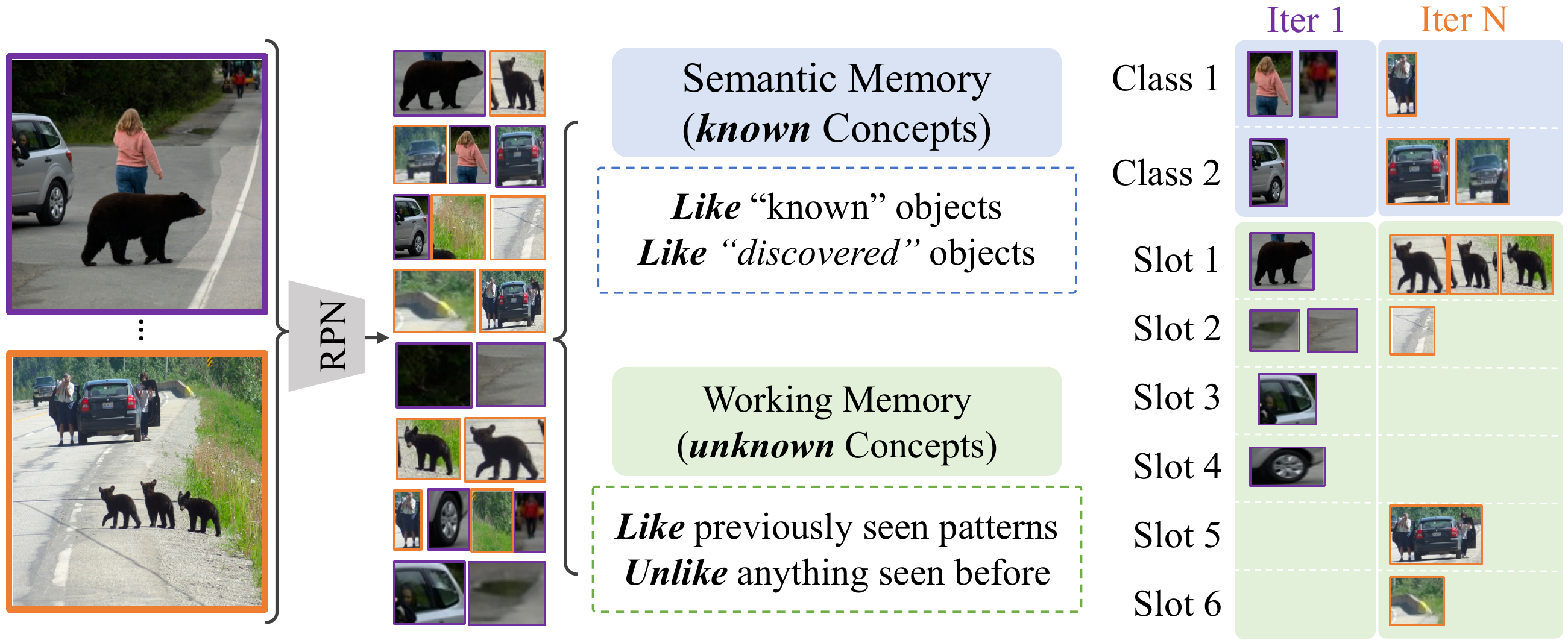}
    \vspace{-0.25in}
    \caption{An example illustrating our dual-memory formulation.}
    \label{fig:explanation}
    \vspace{-0.15in}
\end{figure}

In this section, we illustrate our approach using an example, as shown in Figure~\ref{fig:explanation}. For simplicity, we assume the knowledge of two objects (``known" categories), namely human and car. Our system consists of three main modules, \textbf{\textit{Encoding}}, \textbf{\textit{Storage}} and \textbf{\textit{Retrieval}}. The \textbf{\textit{Encoding}} module is a Region Proposal Network (RPN)~\cite{NIPS2015_5638} that gives candidate regions and corresponding features. The \textbf{\textit{Storage}} module consists of two memory blocks, the \textit{Semantic} and \textit{Working} Memory. The \textit{Semantic} memory consists of slots that store representations to identify regions of ``known" or previously discovered objects while the slots in \textit{Working} memory store representations for potentially ``discoverable" object.  In Figure~\ref{fig:explanation}, the \textit{Semantic} memory is initialized with two slots (human and car classes) and the \textit{Working} memory is null initialized. The \textbf{\textit{Retrieval}} module decides whether a region belongs to the \textit{Semantic} (``known"/discovered) or the \textit{Working} (to be discovered) memory. These modules are discussed in detail in the next section. 

Our system operates sequentially, processing one image at a time. First, the \textbf{\textit{Encoding}} module processes the image and outputs candidate regions and features. The \textbf{\textit{Retrieval}} module assigns each region to either \textit{Semantic} or \textit{Working} memory. In Figure \ref{fig:explanation}, after the first iteration, three regions (two humans and a car) have been assigned to the two slots of the \textit{Semantic} memory and the remaining regions have been assigned to four different slots (Slot 1-4) in the \textit{Working} memory. In the next iteration, the retrieval module assigns three regions (one human and two vehicles) to the semantic memory while, the remaining regions have been assigned to two previously created slots (Slot 1-2) and two new slots (Slot 5-6). Note that the retrieval module can either decide to populate an existing slot in the \textit{Working} memory or create new patterns if necessary. This capability eliminates the need to know the number of ``unknown" objects apriori. Once all the images are processed and we perform \textbf{\textit{Memory Consolidation}}, which step transfers the knowledge acquired in the \textit{Working} memory to the \textit{Semantic} memory, this updating the list of known categories.

\begin{figure*}[t]
    \centering
    \vspace{-0.05in}
    \includegraphics[width=0.9\linewidth]{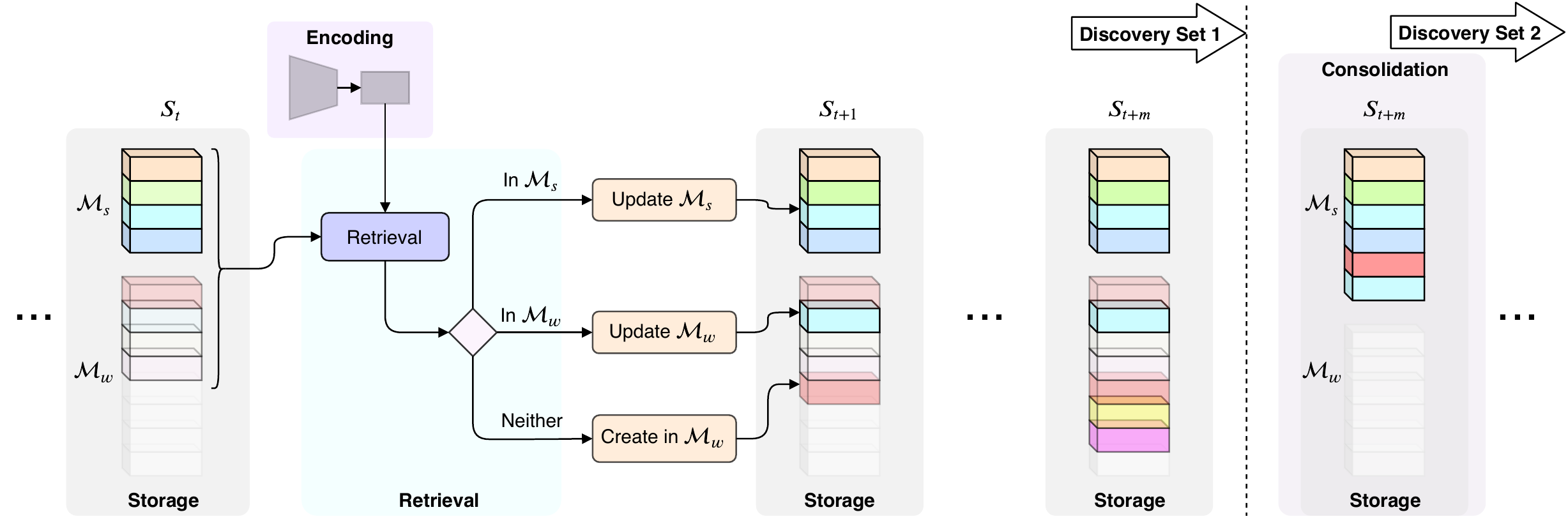}
    \vspace{-0.05in}
    \caption{Overview of our Dual Memory formulation. Refer to Section~\ref{sec:dual_memory} for details.}
    \vspace{-0.2in}
    \label{fig:main}
\end{figure*}

\vspace{-0.05in}
\section{Framework Details}
\label{sec:dual_memory}
\vspace{-0.05in}

In this section, we describe the encoding, storage, and retrieval modules, and how they interact with each other in detail (refer to Figure~\ref{fig:main}).

\noindent\textbf{\underline{Encoding}:}
The goal of the encoding module is to process an input image and extract representations to be subsequently used by the discovery pipeline. Our encoding module is an object detector, Faster R-CNN \cite{NIPS2015_5638}, trained on a dataset (PASCAL VOC \cite{Everingham10}) consisting of a set of known objects (20 objects). Given an image, we use N ($=150$) proposals/boxes from a region proposal network~\cite{NIPS2015_5638} and their corresponding ConvNet features from the classification head (referred to as encoded representation), for subsequent discovery. This module is akin to the encoding process in human memory, which converts sensory inputs to representations to be used by other processes. 
 
\noindent\textbf{\underline{Storage}:}
The storage module consists of two memory blocks: \emph{Semantic} (\ms) and \emph{Working} Memory (\mw). \ms serves the purpose of storing the prior and discovered concepts, and is initialized with `semantic priors', computed using an object detector trained on known classes.  
It resembles the human's long-term memory in that aspect, which stores prior or acquired knowledge. On the other hand, \mw is used to temporarily store and manipulate representations of recently encountered objects, which can potentially be discovered, and is null initialized. The working memory resembles the short-term memory. A concept is considered discovered when we encounter enough instances of it, are able to associate them, and learn a new class from them. One of our key contributions is exploring suitable formulations for this dual memory.

\noindent\textbf{Storage slots.} Both the memory modules are composed of slots, where each slot represents the regions belonging to it from the data encountered so far. A \textbf{slot's representation}, computed using features of all regions belonging to the particular slot, is critical to our approach. If slots represent object classes, then the slot representations are essentially models for these classes and are `trained' using all instances that belong to this class/slot. These representations are used by the retrieval module to decide if a new region/instance `corresponds' to a particular slot. Representing and operating on slots are the cornerstones of our framework, therefore, the slot representation needs to be: (a) effective in representing a large-set of instances corresponding to that slot, (b) efficient in retrieval operations (\eg, associate new instances, updates), and (c) scalable to a large number of object instances and classes. In this work, we explore two choices for slot representations: feature \textbf{centroids} of all the instances belonging to a slot, and a \textbf{classifier} trained on the features, each with a different speed/accuracy trade-off.

The \textbf{centroid representation} is a straightforward centroid of all the instances belonging to a slot, $\mu^\mathbb{C}$; which can be \emph{updated} efficiently using the cumulative moving average. We evaluate the association of an instance feature $f$ with the slot using cosine similarity, $\text{cosine}(f, \mu^\mathbb{C})$.

The \textbf{classifier representation} learns a classifier using all the instances belonging to a slot and classification score is used to evaluate association. We train LDA classifier~\cite{bharat_elda} for each slot.  More specifically, we pose the problem of associating an instance to a slot as a two-class problem, slot \vs background. The LDA classifier is a linear classifier with closed form solution $w = \Sigma^{-1}(\mu^+ -\mu^-)$ and $b = \log(\frac{\pi^+}{\pi^-}) - \frac{1}{2}(\mu^+ - \mu^-)\Sigma^{-1}(\mu^+ + \mu^-)$. Here $\mu^+$ is the mean of all instances in a slot, $\mu^-$ is the mean of background class, $\pi^+$/$\pi^-$ are number of samples in positive/background classes, and $\Sigma$ is the class conditional covariance matrix (assumed to be same for positive/background~\cite{bharat_elda,shrivastavaSA11}). The background class statistics, $\left(\Sigma, \mu_0\right)$, can be estimated efficiently offline using an online update formula operating on boxes from all images from the dataset~\cite{shrivastavaSA11,bharat_elda} (derivations are included in the supplementary material). We use the classifier score, $(w^Tf+b)$, to evaluate the association of the instance feature $f$ to a slot. An added benefit of this closed-form solution is that to \emph{update} the classifier with a new instance, we only need to \emph{update} $\pi^+$.   

Different \emph{slot representations} exhibit different performance and scalability characteristics, and understanding them are crucial for the decision making process of the retrieval module. Centroids are fast to calculate, but not as accurate, whereas classifiers are slow, but accurate. 

Recall that \ms is the stable long term memory, representing classes (known or discovered) with enough samples; and \mw is the volatile short term memory, storing `novel' concepts we encounter, adding/deleting slots frequently, and often with few instances. Therefore, we use classifiers for \ms, where a slot $i$ is $(w_i,b_i)$ or $(\mu_i^+, \pi_i^+, \mu^-, \Sigma, \pi^-)$; and centroids for \mw, where a slot $j$ is $(\mu_j^\mathbb{C}, \pi_j)$ ($\pi_j$ is number of instances, needed for cumulative moving average). Note that since $(\mu^-,  \pi^-, \Sigma)$ are shared across all slots in \ms, the implementation of both \ms and \mw are similar. 
As previously mentioned, \ms is initialized with `semantic priors', which are computed using the detector's outputs on the training set; \ie, we take detections with score $>0.9$ and compute a slot representation for each class in \ms. We limit the total number of slots in \ms and \mw combined to $2000$.

\noindent\textbf{\underline{Retrieval}:}
The retrieval module is tasked with making decisions for our method. 
It takes as input the current state of storage and the features of the region being considered, and makes a decision. For decision making, we draw inspiration from the intuitive item recognition model~\cite{Ratcliff78atheory}. Given a region, we invoke the Semantic Memory $\mathcal{M}_s$ with classifier slot representations for each known (includes previously discovered) objects and use it to determine if the new region corresponds to a known class. As mentioned above, the association with a slot in \ms is evaluated using the classifier score. If so, we update the matching $(\mu_i^+, \pi_i^+)$ in $\mathcal{M}_s$ (`Update \ms' in Figure \ref{fig:main}). If the instance does not correspond to any known object, it likely implies that the region potentially contains a `novel' object. The retrieval process continues by invoking the Working Memory $\mathcal{M}_w$ to check if this region corresponds to a slot in $\mathcal{M}_w$ or a new singleton slot needs to be created (`Update \mw' or `Create in \mw' respectively, in Figure \ref{fig:main}). This is done by computing the cosine similarity between the region's feature and centroids of all the slots in \mw. If a slot in \mw has high similarity with the region, then that slot centroid is updated using a moving average. If the region cannot be associated with any of the slots in \mw, a new slot is created. 

\noindent\textbf{\underline{Memory Consolidation}:}
The Working Memory \mw is responsible for discovering new concepts, and after consistent/repeated occurrence of these concepts, they should be amalgamated with the Semantic memory \ms. Towards this, we propose a memory consolidation step, where representations formed in the Working Memory are added to the Semantic Memory, extending our repertoire of known categories. 
To move slots from \mw to \ms, we train an LDA classifier on the centroids in \mw. A na\"ive consolidation step is to use these new classifiers as slots in \ms. However, since the slots in \mw are essentially online learned clusters, we often encounter fragmented clusters for one concept. To merge multiple clusters of the same categories, we perform an iterative \textbf{merge step} formulated as a graph clustering problem, where each slot is a node and edge weights are based on how well classifiers fire on other slots. We obtain connected components using MinCut and proceed to merge the samples from slots in the same component. This step gives us the flexibility of re-assigning samples of one slot to another based on its relative affinity. Though less fragmented, the resulting slots after the merge operation can be incoherent and lead to semantic drift~\cite{shrivastava_eccv12}. \Eg, samples from a smaller slot (say fire hydrant) can get merged with a larger slot (say bottle) due to their visual affinity to the bottle centroid. To ensure that slots consolidated into \ms are coherent, we perform a final \textbf{refine step}, where we remove all samples from the slot that do not activate its classifier (\ie, $w^{T}x + b < 0$).
After memory consolidation, \mw is reinitialized to its original state (\ie, null), and the number of slots increase in \ms.  
Refer to the supplementary material for detailed merge, refine, and consolidation algorithms.
\begin{table*}[t]
\vspace{-0.2in}
\begin{minipage}[t]{0.32\linewidth}
\setlength{\cmidrulewidth}{0.001em}
\newcolumntype{L}[1]{>{\raggedleft\let\newline\\\arraybackslash\hspace{0pt}\color{gg}}m{#1}<{}}
\renewcommand{\tabcolsep}{4pt}
\renewcommand{\arraystretch}{1.1}
\centering
\caption{\textbf{Large-scale Object discovery} on the \textbf{entire COCO train2014} (80k images). Comparisons with scalable clustering methods using \textbf{AuC} for \emph{unknown} classes.}
\label{tab:final}
\vspace{-.1in}
\footnotesize
\resizebox{0.9\linewidth}{!}{
\begin{tabular}{@{}lccc@{}}
\toprule
\textbf{Method}   &   \textbf{AuC}@0.5 &   \textbf{AuC}@0.2 & \#\textbf{disc.\ objs}\\
\midrule
K-means & 3.34&7.23&42\\
FINCH~\cite{finch} & 3.03& 6.99&42\\
\textbf{Ours} &  \textbf{3.60} & \textbf{9.11} &\textbf{46}\\
\bottomrule
\end{tabular}
}
\centering
\caption{\textbf{Smaller-scale Object discovery} on \textbf{subsets of COCO train2014} (2.5k/20k images). Comparison with contemporary discovery methods using \textbf{AuC} for \emph{unknown} classes.}
\label{tab:final1}
\footnotesize
\renewcommand{\tabcolsep}{4pt}
\renewcommand{\arraystretch}{1.1}
\resizebox{0.9\linewidth}{!}{
\begin{tabular}{@{}lcccc@{}}
\toprule
 \textbf{Method} &\multirow{1}{*}{\makecell[c]{\#\textbf{imgs.}}} &  \textbf{CorLoc} &  \textbf{CorRet} &  \textbf{DetRate} \\
 \midrule 
 OSD~\cite{VoBCHLPP19} & 2.5k&6.62 & 80.00 & 4.73\\
 OSD$^\dagger$ & 2.5k & 6.34 & 70.00 & 5.17 \\
 \textbf{Ours} & 2.5k&43.00 & 64.22 & 48.56\\
\midrule
 rOSD~\cite{Vo20rOSD} & 20k& 15.77 & 100.0 & 11.56\\
 \textbf{Ours} & 20k&41.41 & 64.60 & 46.81\\
\bottomrule
\end{tabular}
}
{\\
\scriptsize$^\dagger$: OSD with ResNet-101 Faster R-CNN proposals and classification-head features (same as \textbf{Ours}).
}
\end{minipage}\hfill
\begin{minipage}[t]{0.65\linewidth}
    \centering
    \strut\vspace*{-\baselineskip}\newline\newline\includegraphics[width=\textwidth]{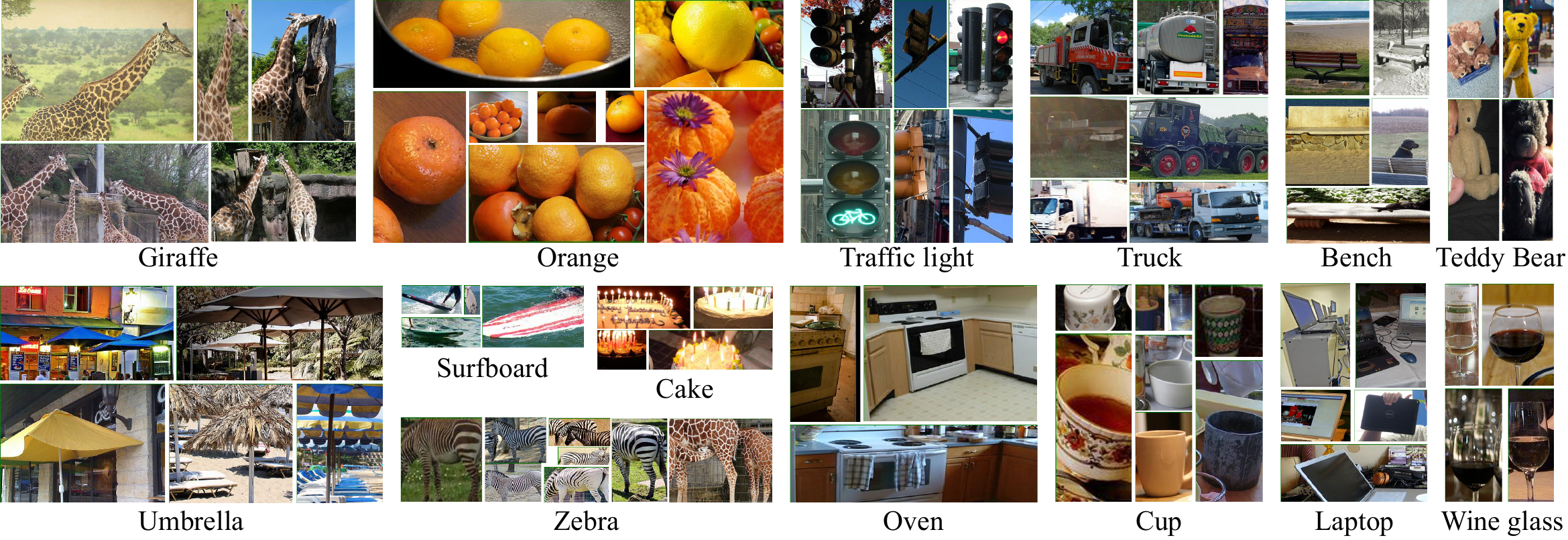}
    \vspace{-0.3in}
    \captionof{figure}{\textbf{Qualitative results (labeled \emph{unknown})} (entire COCO train2014): Concepts discovered by our approach which we can evaluate using the ground-truth annotations for the 60 `unknown' classes.}
    \label{fig:vis}
    
    \strut\vspace*{-\baselineskip}\newline\newline\includegraphics[width=\linewidth]{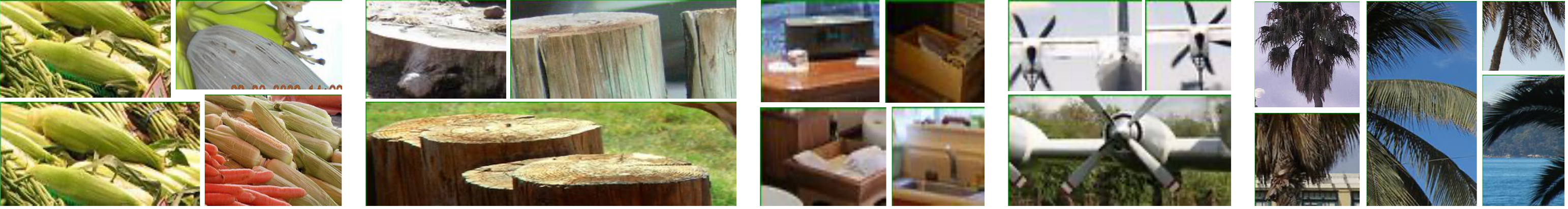}
    \vspace{-0.25in}
    \captionof{figure}{\textbf{Qualitative results (unlabeled \emph{unknown})} (entire COCO train2014): Concepts discovered by our approach which we cannot evaluate since they are unlabeled (not in the 60 `unknown' classes).}
    \label{fig:vis_undiscovered}
\end{minipage}
\vspace{-0.18in}
\end{table*}

\noindent\textbf{\underline{Discovery set}:}
Given a dataset, we perform our discovery process in a sequential manner. This allows us to abstain from making assumptions about the availability of data all at once. To discourage the network from memorizing~\cite{Singh2012DiscPat}, and to leverage the fact that errors made by classifiers on different images will be different~\cite{videossl_cvpr15}, we split the discovery dataset into two disjoint sets (\cf~\cite{Singh2012DiscPat}): \da and \db. First, we run discovery and localization on \da (discovery set) and fill the storage slots. Then, after memory consolidation, we use classifiers in \ms on \db (validation set) to find more examples. The expectation is that the classifiers will find correct and diverse samples, but not detect error modes from the first set \da. After updating \ms using these samples, we swap \da and \db, and continue the process.

\vspace{-0.05in}
\section{Experimental Setup}\label{sec:exp}
\vspace{-0.05in}
\subsection{Benchmark Datasets} 
\vspace{-0.02in}
\noindent\textbf{\underline{Labeled and Discovery Datasets}:} Our benchmark builds on two datasets: (1) Pascal \textbf{VOC} 2007~\cite{Everingham10}, which has 10k images with annotations for 20 classes, and (2) \textbf{COCO} 2014~\cite{lin2014coco} which has $\sim$80k train and 5k validation images with annotations for 80 classes. Our benchmark assumes VOC is the \emph{known} set, with 20 \emph{known} classes, and COCO represents the in-the-wild real-world dataset which we use to benchmark concept discovery and localization. Out of the 80 COCO classes, 60 classes that do not overlap with VOC are treated as \emph{novel} or \emph{unknown} for evaluation. We use ImageNet~\cite{ILSVRC15} for pretraining, but do not use any class similar to the 60 \emph{unknown} classes (discussed later).

\noindent\textbf{Justification and Discussion}. \emph{Difficulty:} Even with 60 unknowns, this is an extremely challenging problem setup, and no other discovery benchmark matches its scale and difficulty; and no current approach can scale to this realistic setup. Another challenge is the different distribution of images -- COCO images are more complex, have more objects per image, and there is a large portion of the discovery dataset which do not have any known class (distractor images). This is also a departure from standard discovery datasets, where all images contain at least one of the objects of interest, providing a strong signal. 
\emph{Semantic drift:} Another benefit is that there are unknown categories (\eg, bear, zebra, truck, laptop) available, which are visually quite similar to the 20 known categories (\eg, dog, horse, car, tv, monitor) from VOC. A discovery and never-ending approach needs to successfully separate these classes to avoid semantic drift~\cite{shrivastava_eccv12}. \emph{Localization focus:} Most importantly, these are detection datasets with bounding box labels, which are suited for evaluating localization. \emph{Unlabeled unknowns:} Note that an approach can certainly discover more than the available 60 unknowns objects. However, due to the lack of labels, we can only evaluate the quantitative performance of part of the discovered and localized objects and visualize other discovered unknowns.

\noindent\textbf{\underline{Pretraining Dataset}:} 
Pretraining on ImageNet 1000 classes, a mainstay in object recognition~\cite{NIPS2015_5638,Girshick:2015,roi8237584, girshick2014rich, shrivastava2016training,LinGGHD20}, creates an issue for any discovery setup -- these 1000 classes might overlap with unknown classes being evaluated. Therefore, to avoid all ambiguity, we identify and eliminate 68 ImageNet classes similar to the 60 \emph{novel} COCO classes; and refer to this split as ImageNet$^{-}$. For the complete list of classes removed from Imagenet for pretraining, training and performance details of the backbone and known detector (Faster-RCNN), we direct the readers to the supplementary material.

\vspace{-0.05in}
\subsection{Evaluation Metrics}
\label{sec:metrics}
\vspace{-0.05in}
Evaluating concept discovery and localization is quite challenging. The two families of approaches discussed in Section~\ref{sec:intro} evaluate on different sets of metrics. For completeness, we report on all metrics used by related approaches, despite their shortcomings, and highlight some metrics that are useful for future works.

\noindent\textbf{\underline{Co-localization/Recall Metrics}:} Several contemporary approaches~\cite{hal-01110036,Vo20rOSD} used three co-segmentation/localization metrics to evaluate concept discovery. \textbf{CorLoc} (correct localization)~\cite{hal-01110036} is defined as the percentage of images in which \emph{any single object} is correctly localized with intersection-over-union (IoU)$>$0.5 with the ground-truth. \textbf{CorRet} (correct retrieval)~\cite{hal-01110036} is defined as the mean percentage of $k$ nearest neighbors ($k$=$10$),  identified by retrieval for each image, that belong to the same (ground-truth) class as the image itself. \textbf{DetRate} (detection rate)~\cite{Vo20rOSD} is the standard recall measure. There are important shortcomings of the metrics described above. CorLoc and {DetRate} measure the localization capabilities and \emph{do not} reflect the discovery performance. Such metrics are more suitable for tasks like co-segmentation/localization (for which they were originally proposed). Neither of these metrics measure the number of objects actually discovered by the algorithm. Moreover, {CorLoc} and {CorRet} assume the number of instances per image is 1, which is overly simplistic and does not hold for datasets like COCO (or even for most VOC images). Finally, DetRate is simply the recall of the region proposal network~\cite{NIPS2015_5638} and does not give any useful information about the discovery algorithm itself. We report these metrics for completeness, but they are not appropriate for our benchmark.
\begin{table*}[t]
\vspace{-0.2in}
\begin{minipage}[t]{0.3\linewidth}
 \setlength{\cmidrulewidth}{0.01em}
\renewcommand{\tabcolsep}{4pt}
\renewcommand{\arraystretch}{1.1}
\centering
\footnotesize
\caption{\textbf{Detection performance (mAP)} for object detectors on COCO minival, trained using \emph{oracle} labels for clusters. $\dagger$: mAP of classes with AP greater than chance.}
\label{tab:final_obj_det}
\vspace{-0.1in}
\resizebox{\linewidth}{!}{
\begin{tabular}{@{}lcccc@{}}
\toprule
 \textbf{Classes (\#)}& \multicolumn{2}{c}{\textbf{GT-IoU}: 0.5} & \multicolumn{2}{c}{\textbf{GT-IoU}: 0.2}\\
 \cmidrule[\cmidrulewidth](l){2-3}
 \cmidrule[\cmidrulewidth](l){4-5}

 & AP@0.5 & AP@0.2&AP@0.5 & AP@0.2 \\
 \midrule
 All (80) & 2.69 & 4.44 & 2.62 & 4.37 \\
 Novel (60) & 1.87 & 3.50 & 1.76 & 3.42\\
 Novel$^\dagger$ & 5.23& 6.47& 5.45& 6.40\\
\bottomrule
\end{tabular}
}
\end{minipage}\hfill
\begin{minipage}[t]{0.68\linewidth}
\centering
\strut\vspace*{-\baselineskip}\newline\newline\includegraphics[width=\linewidth]{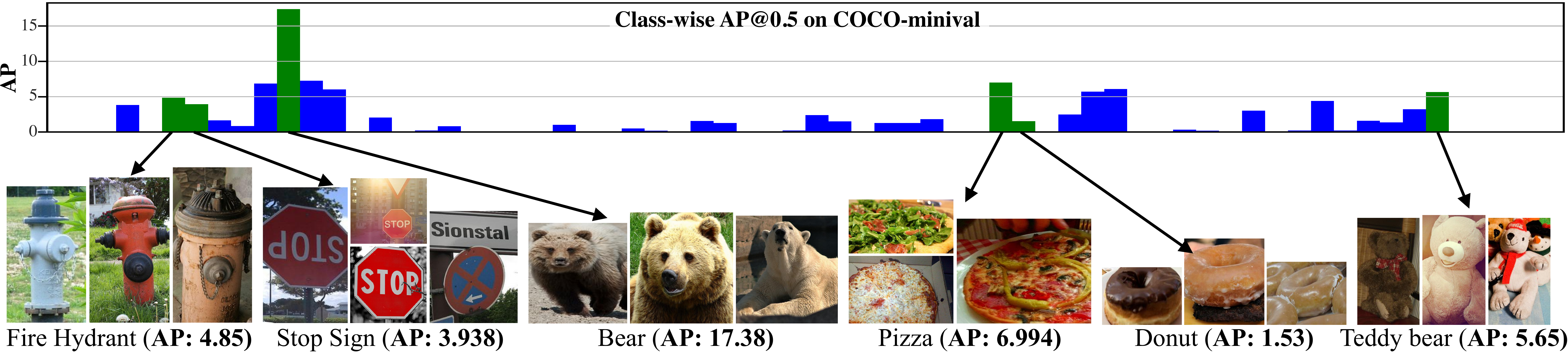}
\vspace{-0.25in}
\captionof{figure}{\textbf{Sample detections and class-wise AP on COCO minival} using our object detectors trained novel classes using \emph{oracle} labels. Many more detection results are provided in the supplementary.}
    \label{fig:det_vis}
\end{minipage}
\vspace{-0.2in}
\end{table*}

\noindent\textbf{\underline{Discovery/Pattern-mining Metrics}:} Some approaches in object discovery~\cite{LeeG11,5540237} and visual pattern mining~\cite{Singh2012DiscPat,doersch2014context} explored forms of Purity-Coverage plots and/or mean Average Prevision (mAP) to evaluate their mining or discovery methods. Following~\cite{LeeG11,5540237,doersch2014context}, we argue that it is natural to evaluate concept discovery methods using such clustering-based metrics as opposed to co-segmentation metrics. We report Area-under-the-curve (\textbf{AuC}) of the Cumulative Purity \vs Coverage plots from~\cite{doersch2014context}.
Along with this, we also report the number of novel objects discovered to evaluate the efficacy of the discovery process. We contend that these metrics together offer a fair assessment across methods. Other plots (\eg, \# of discovered objects \vs coverage, \# of discovered objects/cumulative purity/coverage \vs \# of clusters) offer ablative insights but have limited comparative benefit, and we provide them in the supplementary.

\noindent\textbf{Details.} To compute \emph{cumulative purity}, we first compute purity of all the clusters and sort them in the descending order of their purity. Then, for the $k^\text{th}$ point on the curve, we compute the average purity of top-$k$ clusters and plot it against coverage. We define \emph{purity} of a cluster as $\max_{c \in \mathcal{C}} \frac{\# \text{ of samples assigned to class } c}{\text{Total } \# \text{ of samples in cluster}}$, where $\mathcal{C}$ is the set of all classes in COCO. A sample of a cluster is assigned to class $c$ if it has IoU $\ge$ threshold with ground-truth box (of class $c$) in the image. We define \emph{coverage} as the fraction of ground-truth boxes covered by our method, \ie, if there is at least one proposal with IoU $\ge$ threshold with the ground-truth.

\smallskip\noindent\textbf{\underline{\emph{Oracle} Detection Metric}:}
We also evaluate our approach by training an object detector on the discovered clusters and evaluating it on an held-out set (COCO-minival). We assume the availability of an `oracle,' in the form of ground-truth annotations to assign class labels to each cluster based on majority voting. We use different IoU thresholds for class assignment (0.2/0.5). After training a detector, all boxes rejected by all the classifiers are treated as background. We evaluate using AP at 0.2 and 0.5 IoU. We do not perform any bounding box regression. This provides proxy results, as if we used a human-annotator to provide a single label (out of 60 \emph{unknown} classes) for each cluster. Since these are reproducible `oracle' labels, we encourage future works to continue reporting this metric.

\vspace{-0.06in}
\section{Results and Analysis}
\vspace{-0.05in}

\subsection{Baselines}
\vspace{-0.02in}
\noindent\textbf{\underline{Object discovery baselines}:} We compare our approach with two recent state-of-the-art object discovery methods: OSD~\cite{VoBCHLPP19} and rOSD~\cite{Vo20rOSD}. \textbf{OSD} is an unsupervised method that solves an optimization problem to discover and match object classes among images in a collection. \textbf{rOSD} builds on~\cite{VoBCHLPP19} and is the first large-scale method to perform discovery of multiple objects.

Comparing our approach with other related methods such as~\cite{hal-01153017,hal-01110036,VoBCHLPP19, Vo20rOSD} is not appropriate due several reasons: (1) they perform discovery on VOC, which is eight times smaller in scale compared to COCO, and (2) they do not assume knowledge of known classes. Unfortunately, adapting their official public code to scale to COCO was unsuccessful. \cite{LeeG11,5995527} is closest to our setting, but these methods also suffer from computational scalability and are unable to perform discovery on cluttered environments in COCO.

\noindent\textbf{\underline{Clustering baselines}:} In addition, we compare with two clustering methods: K-means and FINCH~\cite{finch}. \textbf{K-means} requires the number of clusters as input, which is hard to estimate; therefore, we use the number of discovered slots from our approach for K-means for fairness. \textbf{FINCH}~\cite{finch} is a parameter-free clustering method that automatically discovers groupings in the data based on the first nearest neighbors. We pick the clustering that is closest to our discovered number of slots for fairness. Both K-means and FINCH do not scale well to large datasets like COCO with millions of regions. To circumvent this issue, we cluster 150 proposals in 10000 images and do a greedy label propagation to get cluster assignments for all the images~\cite{fergus2009semi}.

\vspace{-0.05in}
\subsection{Object Discovery Results}
\vspace{-0.05in}

\noindent\textbf{\underline{Large-scale Quantitative Evaluation}:} We report the \textbf{AuC} results (\emph{unknown} classes) of our large-scale setup on the COCO2014 train set in Table~\ref{tab:final}, which uses all the 80k images. We recommend this setup for future comparisons. Since no other discovery approach can scale to these many boxes and images, we can only compare our method with K-means and FINCH. Both baselines perform worse compared to our approach, because similarity computation of visual features for clustering is not sufficient to form coherent clusters~\cite{shrivastavaSA11}. Low AuC values across approaches suggest there is tremendous scope for improvement in this domain. Our approach also discovers more objects than these two baselines (46 \vs 42), but there are fourteen categories that are are not discovered.

\begin{table*}[t]
\vspace{-0.2in}
\begin{minipage}[t]{0.32\linewidth}
\centering
\footnotesize
\setlength{\cmidrulewidth}{0.001em}
\newcolumntype{P}[1]{>{\raggedright\arraybackslash}p{#1}}
\renewcommand{\tabcolsep}{2pt}
\renewcommand{\arraystretch}{1.1}
\caption{Ablation analysis for different initializations for \ms performed on mini-train set.}
\vspace{-0.1in}
\label{tab:init}
\footnotesize
\resizebox{\linewidth}{!}{
    \begin{tabular}{@{}P{1.5cm}ccccc@{}}
    \toprule
      \multirow{2}{*}{\makecell[c]{\textbf{Init.}\\(\ms)}} &  \textbf{CorLoc} &  \textbf{CorRet}  &\multicolumn{2}{c}{\textbf{AuC}}& \multirow{2}{*}{\makecell[c]{\#\textbf{disc.}\\\textbf{objs}}}\\
      \cmidrule[\cmidrulewidth](lr){4-5}
     &&&  @0.5 & @0.2 \\
     \midrule 
     $\phi$ (Null) &40.21 &61.31 &2.31&8.00&27\\
      Det.\ Scores  &40.88 & 61.83 &2.97&8.46 & 27\\
     GT Overlap &40.96 & 61.77 & 2.77&8.65 & 30\\
    \bottomrule
    \end{tabular}
}
\end{minipage}\hfill
\begin{minipage}[t]{0.32\linewidth}
\centering
\footnotesize
\setlength{\cmidrulewidth}{0.001em}
\renewcommand{\tabcolsep}{2pt}
\renewcommand{\arraystretch}{1.1}
\newcolumntype{P}[1]{>{\raggedright\arraybackslash}p{#1}}
\caption{Memory consolidation component analysis performed on mini-train set.}
\vspace{-0.1in}
 \label{tab:component}
  \resizebox{\linewidth}{!}{
  \begin{tabular}{@{}P{1.5cm}ccccc@{}}
  \toprule
        \multirow{2}{*}{Method} &  \textbf{CorLoc} &  \textbf{CorRet}  &\multicolumn{2}{c}{\textbf{AuC}}& \multirow{2}{*}{\makecell[c]{\#\textbf{disc.}\\\textbf{objs}}}\\
      \cmidrule[\cmidrulewidth](lr){4-5}
     &&&  @0.5 & @0.2 \\
\midrule
    Na\"ive  & $42.57$& $63.12$ & $3.70$ & $9.60$ &  $44$\\
    + Merge (M)  & $42.57$& $62.21$   & $3.80$ & $9.79$ &  $44$\\
    + M + Refine & $41.20$& $65.88$   & $4.02$ & $10.68$ &  $43$\\
   \bottomrule
  \end{tabular}
  }
\end{minipage}\hfill
 \begin{minipage}[t]{0.29\linewidth}
\centering
\footnotesize
\renewcommand{\tabcolsep}{6pt}
\renewcommand{\arraystretch}{1.1}
\caption{Recall of VOC2007 detectors on the COCO2014 train set.}
\vspace{-0.1in}
\label{tab:recall1}
\resizebox{\linewidth}{!}{
\begin{tabular}{@{}lcccc@{}}
\toprule
 \textbf{Classes (\#)} & \multicolumn{2}{c}{ImageNet} & \multicolumn{2}{c}{ImageNet$^{-}$}\\
  \cmidrule[\cmidrulewidth](l){2-3}
  \cmidrule[\cmidrulewidth](l){4-5}
  & @0.5 & @0.2& @0.5 & @0.2 \\
 \midrule
All (80) & 44.26 & 57.49 & 46.33 & 59.29\\
VOC (20) & 71.26 & 80.07 & 71.67 & 80.47\\
Novel (60) & 35.26 &49.96& 37.88 &52.23\\
\bottomrule
\end{tabular}
}
\end{minipage}
\vspace{-0.18in}
\end{table*}

\noindent\textbf{\underline{Smaller-scale Quantitative Evaluation}:} To compare with~\cite{VoBCHLPP19} (OSD), we modified their code (with some assumptions) to run on 2.5k images from COCO. We run our method on the same split and report the results in Table~\ref{tab:final1}. Originally, OSD~\cite{VoBCHLPP19} used Randomized Prim proposals with whitened HoG features (OSD in Table~\ref{tab:final1}). For fairer comparison, we replace their proposals and features with those used by our approach (ResNet-101+Faster R-CNN) and report the results as OSD$^\dagger$ in Table~\ref{tab:final1}. Our approach handily outperforms OSD on CorLoc and DetRate, whereas OSD performs better on CorRet. In fact, OSD with proposals and features from Faster R-CNN performs worse compared to original OSD, consistent with~\cite{VoBCHLPP19}. One explanation is that deep features are typically trained for image classification while~\cite{VoBCHLPP19} requires region matching. Moreover, since only few regions are localized by OSD, it results in fewer false positives, thereby resulting in a higher CorRet.

Next, we compare with rOSD~\cite{Vo20rOSD}, the first method to scale up to 20k images from COCO (we use the same split from~\cite{Vo20rOSD}) and report the results in Table~\ref{tab:final1} (20k). We observe a similar pattern for each metric as earlier. We would like to reiterate that none of these metrics (CorLoc, CorRet, and DetRate) adequately measure the object discovery performance and are only provided for completeness.

\noindent\textbf{\underline{Qualitative Results}:} We show some discovered clusters in Figure~\ref{fig:vis} from the COCO train set which we evaluated in this section (60 labeled \emph{unknown} in COCO). In Figure~\ref{fig:vis_undiscovered}, we show some discovered clusters which we could not evaluate because they are unlabeled in COCO (unlabeled \emph{unknown}). Many more examples are provided in the supplementary.

\vspace{-0.05in}
\subsection{Object Detection Results} 
\vspace{-0.05in}
To demonstrate the performance of our approach on novel data, we evaluate detectors obtained from our approach on COCO-minival using the \emph{Oracle} Detection Metric methodology described in $\S$\ref{sec:metrics}. Note that this is a purposeful benefit of our VOC-COCO setup, where we can assume the availability of an `oracle' for the 60 \emph{unknown} classes and train a detector from our discovered clusters. In Table~\ref{tab:final_obj_det}, we report results by assigning labels with different ground-truth IoU thresholds (0.2/0.5) and evaluating COCO AP at 0.2 and 0.5 IoU. We report AP results for all 80 COCO objects, all 60 novel objects, and for novel objects whose performance is greater than chance (Novel$^\dagger$). Novel$^\dagger$ objects achieve $>$5 AP performance, which is reminiscent of early days of complex object detection.

Next, we show the per-class AP, for the model evaluated at 0.5 IoU for the discovered classes in Figure~\ref{fig:det_vis}. We also visualize the detections on COCO-minival for few random categories. Evidently, the detectors display a lot of intra-class variation. We achieve the highest AP of $\sim 17.38\%$ for the bear class and a lowest mAP of $0.08\%$ for traffic lights.

\vspace{-0.05in}
\subsection{Ablation Analysis}
\vspace{-0.05in}
Finally, we evaluate several design choices in our framework. For tractability, we perform all ablation studies on a randomly selected subset of $5000$ training images from the COCO2014 train set. To ensure that results from the analysis translate to the entire dataset, we compare the distribution of objects in this mini-train and the entire train set in the supplementary.

\noindent\textbf{\underline{Semantic Memory Initialization}:} To understand how the initialization of Semantic Memory \ms with features of known categories influences the discovery process, we run the discovery method without this initialization and report results in Table~\ref{tab:init} (rows 1-2), which demonstrates the importance of prior knowledge. We use detection results to initialize \ms (see  $\S$\ref{sec:dual_memory}). Here, we discuss an alternate, higher-quality initialization, where we assume access to ground-truth annotations for the 20 known classes in COCO (which might not be available in most scenarios). For each known category, we compute the feature centroid of all boxes with IoU$>0.5$ with ground-truth. Results in Table~\ref{tab:init} (row 3) shows that this helps discover more objects.

\noindent\textbf{\underline{Memory Consolidation Component Analysis}:} We analyze the contribution of different components of our memory consolidation step in Table~\ref{tab:component}. For evaluation, all clusters with regions from less than $5$ images are dropped. The merge step improves the AuC for \emph{unknown} classes because it reassigns cluster memberships based on affinity scores. The refine step further improves the purity of a cluster by dropping all incoherent samples, which increases AuC and CorRet, but drops CorLoc slightly.

\noindent\textbf{\underline{Recall Analysis}:} Since most discovery approaches, including ours, rely on region proposals, we explore how good are VOC-trained proposal generators for unseen `novel' classes. We compute the recall of these proposals at IoU thresholds of $0.2$ and $0.5$ on the entire COCO2014 train split and report results in Table~\ref{tab:recall1} for all classes (`All'), for 20 known VOC classes (`VOC'), and for 60 unknown classes (`Novel'). As we can see, the recall for unknown classes significantly lags behind known classes. This gives us a performance upper-bound for all discovery methods discussed in this work. This also strongly suggests that future research in better proposal generators is needed that can generalize to unseen categories.

\noindent\textbf{\underline{Discovery Rounds}:} Since our approach can run in a never-ending fashion, we assess how many rounds are needed for convergence and how the performance changes across iterations. These results are reported in the supplementary.

\vspace{-0.1in}
\paragraph{Conclusion and Future Work.}
\vspace{-0.1in}
We presented a dual memory formulation, which can exploit prior knowledge about known objects to discover and localize novel categories in-the-wild. We perform extensive experiments to validate our claims. However, as the raw numbers indicate, there is a lot of scope for improving current object discovery and localization methods, especially for complex scenes and realistic benchmarks. One immediate future work is to adapt deep neural network-based region proposal methods to generalize beyond their training datasets and seen classes. Another exciting future direction is a new paradigm of supervised recognition, which is more suited for adaptation to in-the-wild discovery setups.

\smallskip
\noindent\textbf{Acknowledgement.} This project was partially supported by DARPA SemaFor (HR001119S0085) and DARPA SAIL-ON (W911NF2020009), and Amazon Research Award to AS. We thank Pulkit Kumar, Alex Hanson, Max Ehrlich, and the reviewers for valuable feedback.

{\small
\bibliographystyle{ieee_fullname}
\bibliography{egbib}
}
\end{document}